\documentclass[10pt,twocolumn,letterpaper]{article}

\usepackage{cvpr}              
\usepackage{booktabs}
\usepackage{algorithm}
\usepackage{algpseudocode}
\definecolor{cvprblue}{rgb}{0.21,0.49,0.74}
\usepackage[pagebackref,breaklinks,colorlinks,allcolors=cvprblue]{hyperref}

\title{PanoDreamer: Consistent Text to 360-Degree Scene Generation}

\author{%
  Zhexiao Xiong$^{1,2}$\thanks{Major work was done during internship at OPPO US Research Center.}\quad 
  Zhang Chen$^{1}$\quad  
  Zhong Li$^{1}$\quad  
  Yi Xu$^{1}$\quad  
  Nathan Jacobs$^{2}$\\[0.5ex]
  {\normalsize $^1$ OPPO US Research Center}\quad {\normalsize $^2$ Washington University in St. Louis}
}
\begin{document}
\maketitle

\begin{abstract}

Automatically generating a complete 3D scene from a text description, a reference image, or both has significant applications in fields like virtual reality and gaming. However, current methods often generate low-quality textures and inconsistent 3D structures. This is especially true when extrapolating significantly beyond the field of view of the reference image. To address these challenges, we propose PanoDreamer, a novel framework for consistent, 3D scene generation with flexible text and image control. Our approach employs a large language model and a warp-refine pipeline, first generating an initial set of images and then compositing them into a 360-degree panorama. This panorama is then lifted into 3D to form an initial point cloud. We then use several approaches to generate additional images, from different viewpoints, that are consistent with the initial point cloud and expand/refine the initial point cloud. Given the resulting set of images, we utilize 3D Gaussian Splatting to create the final 3D scene, which can then be rendered from different viewpoints. Experiments demonstrate the effectiveness of PanoDreamer in generating high-quality, geometrically consistent 3D scenes.

\end{abstract}

\section{Introduction}

The immense potential of text-to-3D applications in VR/AR platforms, industrial design, and the gaming industry has driven substantial research efforts toward establishing a robust approach for immersive scene content creation. Recent developments in diffusion models~\citep{ruiz2023dreambooth,yuan2023customnet,li2024blip} make it possible to generate high-quality, geometrically-correct images from text, allowing for customized 2D content generation.

Based on the recent advance in 2D text-to-image generation~\citep{podell2023sdxl,xiong2024groundingbooth,qiao2025genstereo,zhang2023adding}, many works have begun focusing on 3D scene generation. Some works~\citep{chen2023scenedreamer,chung2023luciddreamer,yu2024wonderjourney} first generate an initial point cloud based on a reference image, employing a progressive warp-and-refine approach to complete the 3D scene reconstruction. However, due to the limited camera field-of-view (FoV), these approaches require multiple iterations to generate a complete scene, with each iteration relying solely on information from the previous stage. As a result, error accumulation from monocular depth estimation and artifacts from diffusion generation hinder these models' ability to maintain long-term geometric and appearance consistency, particularly with large camera movements.

To overcome these challenges, recent works have leveraged panorama-to-3D scene generation~\citep{perf2023,zhou2025dreamscene360} to generate scenes with a larger FoV. Utilizing advancements in text-to-panorama generation~\citep{panfusion2024}, these methods use panoramas as intermediate representations of the 3D scene, subsequently obtaining 3D representations using Neural Radiance Fields (NeRF) or 3D Gaussian Splatting (3D-GS). However, since the geometry is based on a single panorama, the generated 3D scenes have a limited spatial extent and are significantly impacted by occlusions. As a result, users have limited freedom to move about the scene, greatly limiting the usefulness of the 3D model.

In this work, we propose PanoDreamer, a novel framework that enables global-level scene generation with geometric consistency and allows for customized 3D scene extension. Our approach adopts a multi-stage pipeline: first generating a static panoramic scene, followed by extending the scene dynamically based on user-defined initial images and camera trajectories. To generate the static panoramic scene, given a text prompt and/or a user-provided reference image, we synthesize images from an initial viewpoint using an LLM engine and composite them into a complete equirectangular panorama. This panorama is then lifted into 3D to create an initial point cloud. We then
generate a set of additional images from different viewpoints. We use a view-conditioned video diffusion model to generate sequences based on user-specified initial images and trajectories, enabling both continuous, geometrically consistent scene generation and flexible control over viewpoint shifts.

The resulting point clouds are composed into a global point cloud using depth alignment, followed by 3D Gaussian Splatting to produce the 3D scene representation. To enhance the scene completeness, we propose a strategy to generate a set of supplementary views and employ a semantic-preserving generative warping framework to inpaint occluded regions. These supplementary viewpoints, along with their inpainted images, are used to refine the 3D Gaussians, thereby reducing artifacts and enhancing scene completeness.

Our main contributions can be summarized as follows:
\begin{itemize}

\item We propose PanoDreamer, a holistic text to 360-degree scene generation pipeline, which achieves consistent text-to-360-degree scene generation with customized trajectory-guided 
scene extension.

\item We introduce semantically guided novel view synthesis into the refinement of 3D-GS optimization, reducing artifacts and improving geometric consistency.

\item Experiments show the effectiveness of our model in generating geometrically consistent and high-quality 360-degree scenes.

\end{itemize}

\section{Related Work}

\paragraph{Panorama Generation}
With the development of diffusion models~\cite{rombach2022high}, many studies have sought to generate panoramic scenes using existing text-to-image diffusion techniques~\citep{bar2023multidiffusion, Tang2023mvdiffusion,cai2024magic}. These methods often use text-to-image generation models or image outpainting techniques to first synthesize multi-view images, subsequently generating the panorama through equirectangular projection. MVDiffusion~\cite{Tang2023mvdiffusion} proposed a correspondence-aware attention to generate text-conditioned multi-view images or extrapolates one perspective image to a full 360-degree view. Some later methods finetune the diffusion models to generate panoramas. StitchDiffusion~\cite{wang2024customizing} used Low-Rank Adaptation (LORA)~\citep{hu2021lora} to generate panoramic images and achieves customized generation, PanFusion~\cite{zhang2024taming} tried to use panoramic diffusion in the latent space, Diffusion360~\citep{feng2023diffusion360} utilized DreamBooth~\citep{ruiz2023dreambooth} finetuning alongside circular blending to produce panoramas in both text-to-panorama and single-image-to-panorama tasks, and MVPS~\citep{xiong2024mixed} used geospatial information to guide panorama generation. Our method supports both text-only input and combined text and image conditions, achieving flexible panorama generation.

\paragraph{Conditional Video Diffusion Models}
With the increasing need for multi-modality control, controlled video generation have evolve rapidly. Benefiting from previous customized image generation methods~\citep{zhang2023adding,li2023gligen,xiong2024groundingbooth}, conditional video diffusion models also allow for multiple control guidance, including text~\citep{chen2023videocrafter1,chen2024videocrafter2,singer2022make,xing2025dynamicrafter}, RGB images, depth~\citep{esser2023structure}, semantic maps~\citep{peruzzo2024vase} and trajectory~\citep{niu2024mofa,yin2023dragnuwa}. Recent studies regard video diffusion models as a strong tool in downstream tasks, such as stylization~\cite{liu2023stylecrafter}, motion control~\cite{wang2024videocomposer}, novel view synthesis~\citep{yu2024viewcrafter}. Specifically, view conditioned video diffusion models such as ViewCrafter~\cite{yu2024viewcrafter} regarded point cloud renders as control to synthesis novel view generic scenes either from both single or sparse images, which enables a point cloud completion and benefit the downstream tasks. Our work leverage ViewCrafter's generalization ability to help customize the extention of scene generation with geometric consistency.

\paragraph{Dreaming-based 3D Scene Generation}
Synsin~\cite{wiles2020synsin} was one of the pioneering methods that employed a warp-and-refine strategy to generate point clouds of a scene. With the rapid advancement of diffusion models and Score Distillation Sampling (SDS)-based techniques, dreaming-based text-to-3D generation has emerged as a popular approach for creating 3D content. Early methods predominantly utilized Neural Radiance Fields (NeRF)\citep{zhang2023text2nerf,mildenhall2020nerf} or mesh-based representations\citep{song2023roomdreamer,hoellein2023text2room} to achieve scene reconstructions. More recent works, such as LucidDreamer~\citep{chung2023luciddreamer}, have employed 3D Gaussian Splatting~\cite{kerbl3Dgaussians} to achieve consistent rendering with greater geometric flexibility. However, these approaches primarily focus on generating forward-facing scenes, which restricts the scalability for more extensive scene generation.

To achieve global-level 360-degree scene generation, some recent works have turned to using panoramic representations as an intermediate stage for comprehensive scene synthesis. PERF~\cite{perf2023} was the first to propose a 360-degree novel view synthesis method by training a panoramic neural radiance field from a single panorama, enabling free movement within a 3D environment. Despite its potential, this approach remains largely confined to indoor scenes. Later methods like DreamScene360~\cite{zhou2025dreamscene360} and HoloDreamer~\cite{zhou2024holodreamer} advanced the concept by adopting panoramic Gaussian splatting for 360-degree scene generation. Additionally, approaches such as LayerPano3D~\cite{yang2024layerpano3d} introduced layered panorama generation techniques to manage complex scenes, subsequently lifting these layers into 3D Gaussian splatting representations.

In our work, we propose a method for text-to-360-degree scene generation that also utilizes panoramas as an intermediate representation, combined with 3D Gaussian splatting to generate the final 3D scene. Our approach not only overcomes the limitations of previous methods but also enhances geometric consistency and provides greater flexibility in scene extension and customization.

\begin{figure*}[t!]
    \centering
    \includegraphics[width=0.95
    \linewidth]{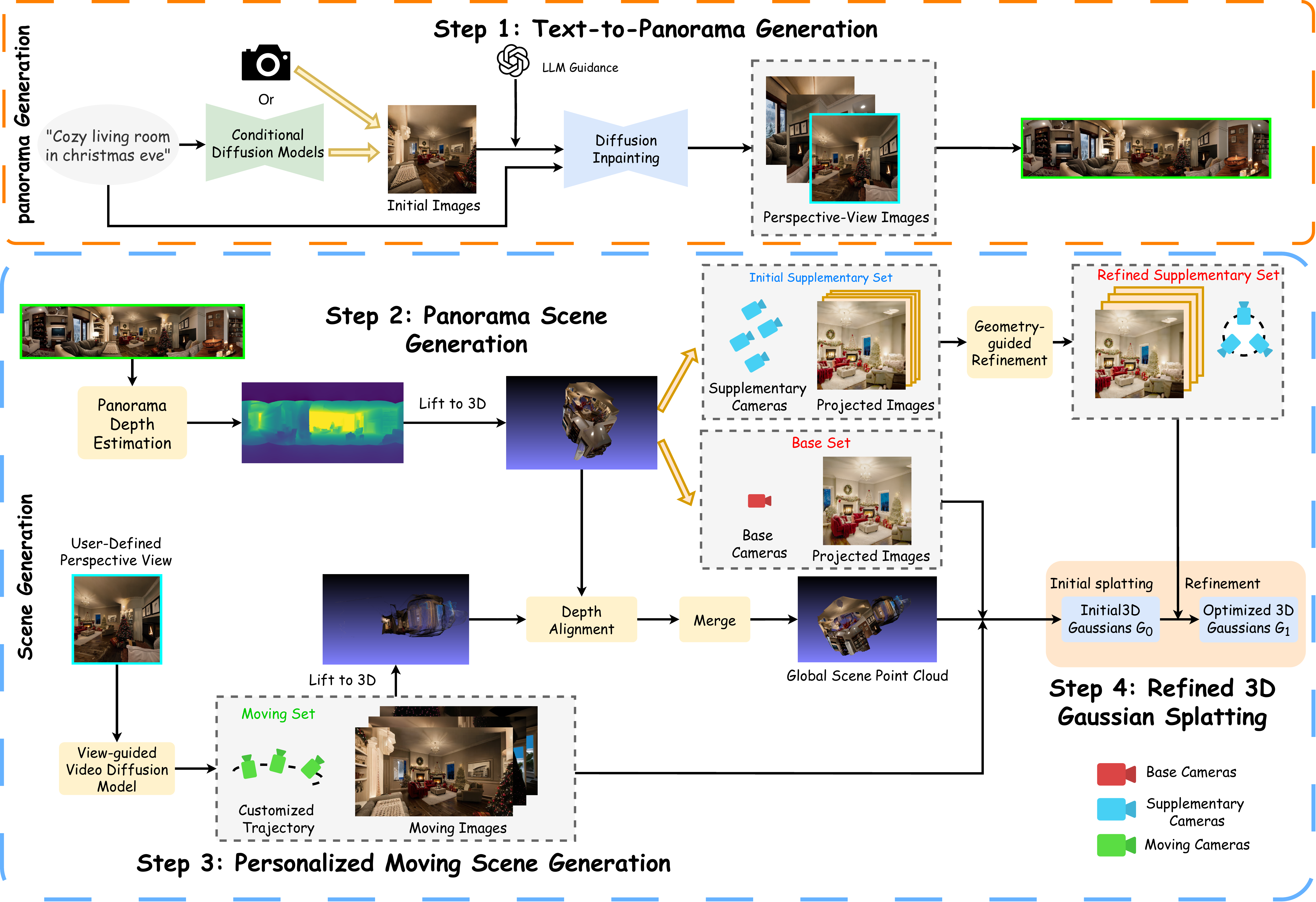}
     \caption{Modules of our proposed framework. (a) Text-to-Panorama Generation: we use LLM as guidance to guide the generation of perspective-view images. (b) Scene Generation: we divide it into static panorama scene generation and customized moving scene generation. Besides base camera set, we compose an additional supplementary camera set for the static panorama scene and use semantic-preserved warping to generate the missing region, which is used for 3d gaussian splatting refinement.}
    \label{fig:warp}
\end{figure*}

\section{Preliminary}
\paragraph{3D Gaussian Splatting.}
3D Gaussian Splatting (3DGS)~\citep{kerbl3Dgaussians} is a recent pioneer method for novel view synthesis and 3D reconstruction,utilizing the multiview calibrated images from Structure-from-Motion. Unlike implicit representation methods like NeRF~\citep{mildenhall2021nerf}, 3D-GS renders in an explicit manner through splatting, achieving real-time rendering and reduced memory consumption. The 3D Gaussians can be queried as:
\begin{equation}
G(x)=e^{-\frac{1}{2}(x)^T \Sigma^{-1}(x)},
\end{equation}
where $x$ represents the distance between the center position $\mu$ and the query point. During the rendering process, the color of $r$ on the image plane is rendered sequentially with point-based volume rendering technique through:
\begin{equation}
C(r)=\sum_{i \in \mathcal{N}} c_i \sigma_i \prod_{j=1}^{i-1}\left(1-\sigma_j\right), \quad \sigma_i=\alpha_i G\left(x_i\right),
\end{equation}
where $\mathcal{N}$ represents the number of sample points on the ray
$r$, $c_i$ and $\alpha_i$ denote the color and opacity of the $i$-th Gaussian, and $x_i$ is the distance between the point and the $i$-th Gaussian.

\section{Method}
Given a text prompt $T$ or an optional user-provided inintial image $I$, we aim to get 3D scene-representation with global-level consistency, and allows for free camera movements among different sub-scenes. In this section, we first introduce the static text-to 360-degree panorama generation in Sec.~\ref{3.1}, then introduce the panorama scene generation and supplementary moving scene generation in Sec.~\ref{3.2} and Sec.~\ref{3.3} respectively. Finally we introduce the scene generation with 2-stage 3D gaussian splatting in Sec.~\ref{3.4}.

\subsection{Text-to-Panorama Generation}\label{3.1}
Traditional text-to-panorama generation methods often rely on a single prompt, restricting their ability to generate panorama with rich content and may lead to repeated content. Inspired by L-Magic~\cite{cai2024magic}, we decompose the panorama generation process into two distinct stages: LLM-guided perspective image generation and panorama composition. First, an image is projected into the unit sphere by defining the vertices $V$ on each image pixel and creating edges between adjacent pixels. Then we use a warp-and-inpainting strategy to get novel view images. Specifically, given a rotation matrix $R$ from viewpoint $A$ to viewpoint $B$, the pixel in the other frame is computed as $\mathbf{P}_{\mathrm{rot}}^{\mathrm{i}}=\mathbf{R} \mathbf{P}_{\mathrm{0}}$, where the camerea field of view(FoV) is set to be 100 degrees. In this process, binary mask $M$ is used to ensure that the inpainting is constrained in the non-overlapping region between adjacent views. In the inpainting stage, we use Stable Diffusion V2 inpainting model~\cite{rombach2022high} to extrapolate the large missing region of the warped view. To effectively remove the artifacts in the inpainting stage, we use use GPT4-o to achieve instruction-guided inpainting, which helps remove the duplicated objects.

To remove the blurry region in the border of the perspective view images and enhance the detail, we use diffusion-based super-resolution~\citep{wang2024sinsr} to increase the resolution of each perspective-view image from $512\times 512$ to $2048 \times 2048$. We warp the high-resolution image to a lower-resolution next-view image and use super-resolution again. This iterative process is repeated to obtain high-resolution images for all perspective views.

Finally, equirectangular projection is used to warp all perspective views to the same equirectangular plane and merge into a seamless panorama. Since the merged panorama does not cover the full 180-degree FoV in the vertical direction, we utilize a panorama inpainting method~\citep{wu2023panodiffusion} to fill in the missing regions at the top and bottom. Eventually, we can get a complete equirectangular panorama with 180-degree vertical FoV.

\subsection{Panorama Scene Generation} \label{3.2}
After synthesizing the panorama, we use a zero-shot panorama depth estimation network~\citep{wang2024depth} to get the depth of the panorama, based on which we lift the panorama to 3D and get the point cloud of the panorama. As the lifted point cloud only contains point clouds generated from the same location, artifacts exists due to the occlusions. In the boundary area of the objects, the point cloud is not continuous, leading to holes in the projected image when translating camera positions. Therefore, we propose a diffusion-based refine method to refine the projected images. 

Previous scene generation approaches often employed a warp-and-refine strategy to progressively generate the scene. However, this strategy heavily relies on the accuracy of monocular depth estimation, often leading to blurry results in complex scenes with noisy depth maps. Moreover, semantic details are frequently lost, particularly during large viewpoint changes. A recent work, GenWarp~\cite{seo2024genwarp}, has demonstrated effective geometric and semantic-preserved warping by augmenting cross-view attention with self-attention in the diffusion process. Based on this, we use a geometry-preserved warping method to fill in missing points in the point cloud. 
Specifically, we first construct a base camera set $P_B={p_1, p_2, \cdots, p_m}$, where each camera is positioned at the center of the point cloud and uniformly faces different directions across 360-degree. We then project images from these cameras to form the base image set $I_B ={ I_1, I_2, \cdots, I_m}$. Since the cameras are positioned at the center of the point cloud, these base images are largely free from artifacts. Subsequently, we sample an additional $4m$ supplementary cameras $C_S$, all sharing the same intrinsics $K$. For each base camera $c_m$, we translate it up, down, left, and right by a uniform offset, forming a supplementary camera set $P_S = {p_{(m,1)}, p_{(m,2)}, p_{(m,3)}, p_{(m,4)}}$.

Based on the base image $I_m$ and the corresponding depth map $D_m$ obtained via multiview depth estimation with Open3D, we conduct the semantic-preserved generative warping in the latent space through GenWarp, represented as:
\begin{equation}
E_{m,n} = \operatorname{GenWarp}\left(E_m; D_m, P_{m \rightarrow (m,n)}, K\right),
\end{equation}
where $E_m$ and $E_{m,n}$ represent the Fourier features of the base image and the projected image from the supplementary camera respectively, $P_{{m} \rightarrow (m,n)}$ denotes the relative camera pose from the base camera to the corresponding supplementary camera.
The resulting feature embedding is then decoded to yield the inpainted supplementary image $I_{m,n}$. We also get the occlusion mask $M_{m,n}$, represented as the supplementary mask set $M_S$, which is also shown in Fig.~\ref{fig:warp}. Therefore, we get the supplementary camera set $P_S$ and the corresponding supplementary image set $I_S$ based on semantic-preserved generative warping. We save $I_B$, $P_B$, $I_S$, $P_S$ and $M_S$ for the subsequent 3D-GS optimization.

\begin{figure*}[t!]
    \centering
    \includegraphics[width=\linewidth]{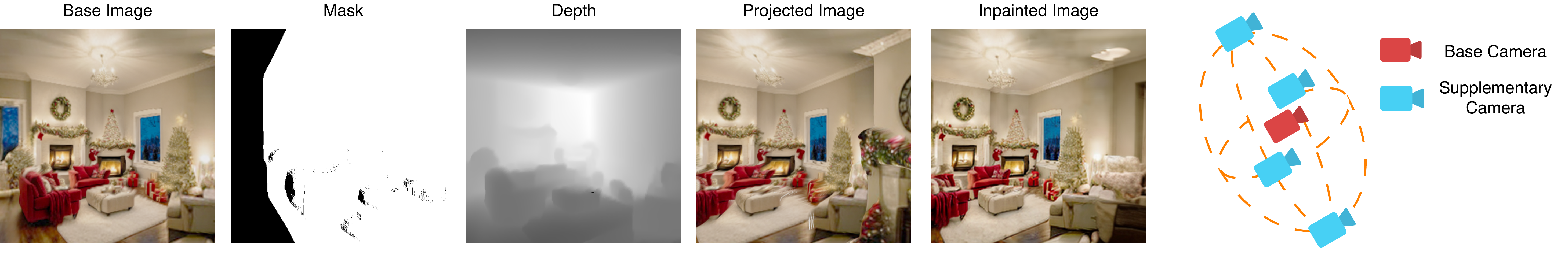}
    \caption{Semantic-preserved Refinement: For each base camera, we apply supplementary cameras to up, down, left, and right directions respectively. For each supplementary camera, we get projected images through semantic-preserved generative warping~\citep{seo2024genwarp} to fill the missing area brought by occlusion. }
    \label{fig:submodules}
\end{figure*}

\subsection{Supplementary Moving Scene Generation}\label{3.3}
Although the panorama point cloud has a 360-degree view of the scene, it is often limited to a single position and restricted by occlusions such as walls and furniture, especially for indoor scenes like a single room. In practical applications, users may expect to navigate to other areas to obtain more comprehensive, global views of the environment. To enable this, we utilize view-guided video diffusion models~\citep{yu2024viewcrafter} to generate moving scenes as supplementary views. Users can select a perspective-view image generated during the text-to-panorama process as the initial reference and define a custom camera trajectory.

Specifically, based on the initial image and the user-defined camera trajectory, we render a sequence of video frames along a target direction outside the panorama scene. Leveraging the reconstruction capability of the latest dense stereo methods, such as DUSt3R~\citep{wang2024dust3r}, we lift the moving scenes into a 3D point cloud, while also capturing the camera poses for each frame in the world coordinate system. However, directly aligning the moving scenes with the panorama scene results in two challenges: (1) overlapping regions between scenes, and (2) depth misalignment due to differences in the depth estimation methods used for the panorama and the moving scenes. 

To mitigate these issues, we implement a masking strategy that eliminates duplicate regions by excluding pixels corresponding to the initial frame from the point cloud. Given the initial image $I_0$ and a specified trajectory, we generate $m$ frames of images $I_1, I_2, \cdots, I_m$ using view-guided video diffusion models. We then uniformly sample $n$ images from the video sequence and perform sparse-view reconstruction with DUSt3R.

In the selected set of images, the first frame overlaps with the initial image, and we ensure that all the camera poses of the frames are in the same world coordinate system. To address the overlapping regions between the scenes, we create a view mask $M_0$ for the points that lie within the view of the first camera.
We first transfer the point cloud from the world coordinate to the camera coordinate, represented as:
\begin{equation}
\mathbf{P}_{c a m}=\mathbf{T}_{w 2 c} \cdot \mathbf{P}_h^T ,
\end{equation}
where $\mathbf{P}_{c a m} \in \mathbb{R}^{512 \times 1024 \times 4}$. For the point cloud in the camera coordinate system, we check if the point is in the camera view through $M_{front}=\left(\mathbf{P}_{\text {cam }}[\ldots, 2]>0\right)$, Next, we project the 3D points in the camera frame onto the image plane using the camera's intrinsic matrix $\mathbf{K}$ through: 
\begin{equation}
\mathbf{u}_h=\mathbf{K} \cdot \mathbf{P}_{c a m}[\ldots,: 3]^T,
\end{equation}
which gives us the homogeneous coordinates in the image plane. We then normalize these to obtain the pixel coordinates through:
\begin{equation}
\mathbf{u}=\frac{\mathbf{u}_h[\ldots, 0]}{\mathbf{u}_h[\ldots, 2]}, \quad \mathbf{v}=\frac{\mathbf{u}_h[\ldots, 1]}{\mathbf{u}_h[\ldots, 2]},
\end{equation}
where $\mathbf{u}$ and $\mathbf{v}$ are the horizontal and vertical pixel coordinates on the image plane, indicating the column and row position of the 3D point when projected onto the camera image, respectively. 
The overall mask $\mathbf{M}$ is then computed as:
\begin{equation}
\mathbf{M} = \neg{(\mathbf{M}_{bound} \cap \mathbf{M}_{front})},
\end{equation}
which indicates the indices of the points that are not within the first camera view. We apply this mask to retain only the points that are not visible in the first camera's view, represented as:
\begin{equation}
\mathbf{P}_{\text {f}}= \begin{cases}\mathbf{P}, & \text {if M }=1 \\ 0, & \text { otherwise }\end{cases},
\end{equation}
and we get the filtered point cloud $\mathbf{P}_{\text {f}}$ of the moving scene. At the final stage of generation, we remove the overlapping regions of the other frames from the first frame's view, obtaining a point cloud of the moving scene that is non-overlapping with the initial point cloud.

To keep the depth consistency between the moving scene and the static panorama scene, we use a depth scaler optimization strategy to maintain the depth consistency between the point cloud of the panorama scene and the moving scene. Specifically, we first convert the depth values into disparity, and then use a least squares-based optimization strategy to minimize the disparity difference between the first frame of the view-controlled video generated by DUSt3R and the corresponding disparity of the initial image obtained from the panorama depth estimation method. The optimization is represented as:
\begin{equation}
\min _{\alpha, \beta}\left\|\mathbf{M} \odot\left(\frac{\alpha}{d_p}+\beta-\frac{1}{d}\right)\right\|^2,
\end{equation}
where $\alpha$ and $\beta$ represent the scale and shift factors respectively. The mask $\mathbf{M}$ ensures that the depth alignment is conducted only in the overlapping regions, $d$ represents the depth of the initial image of the moving scene and $d_p$ represents the depth of the panorama in the corresponding region. The aligned depth is then given by:
\begin{equation}
\hat{d}=\left(\frac{\alpha}{\hat{d}_p}+\beta\right)^{-1},
\end{equation}
where $\hat{d}$ is the rectified depth of the initial image of the moving scene. Then we get the scale factor $\gamma$ through $\gamma = \frac{\hat{d}}{d}$. Finally, we apply a $7 \times 7$ Gaussian kernel to smooth $\hat{d}$ at the mask edges, ensuring seamless transitions. For the depth of each subsequent frame $d_i$, we multiply with the same scale factor $\gamma$ to get the rectified depth $\hat{d_i}$, ensuring that the point cloud of the moving scenes remain consistent in scale with the panorama point cloud.

Defined by users, the area that need to be expanded from the moving scene can be repeated, and we get the moving scenes ${P_M^1, P_M^2, \cdots, P_M^i}$. Finally, As both the static panorama scene and the moving scene are in the world coordinate system, we fuse the point cloud of the moving scenes with the panorama scene to get the final global-level point cloud, represented as:
\begin{equation}
\Omega = \bigcup_{i=1}^{|P|} \varphi\left\{P_0, P_M^1, \cdots, P_M^{i}\right\},
\end{equation}
where $\varphi$ represents the depth alignment function, $P_0$ represents the initial panorama point cloud, $M_i$ represents the $i-th$ moving scene point cloud, and $\Omega$ represents the complete global-level point cloud. 

\begin{figure*}[t!]
    \centering
    \includegraphics[width=\linewidth]{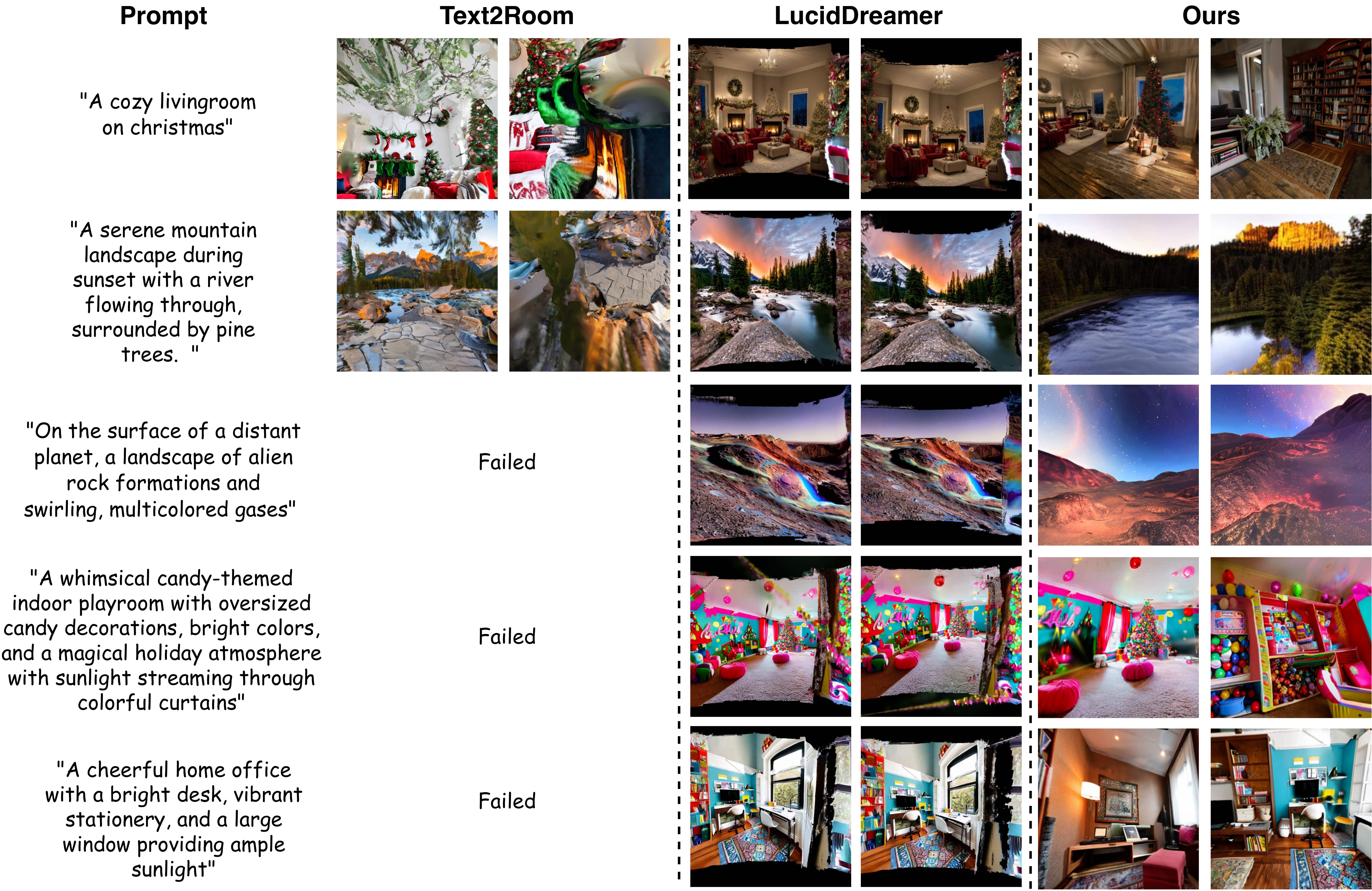}
    \caption{Comparison of results. For Text2Room the images are projected from mesh and for LucidDreamer and our method, the images are projected from 3D Gaussians. Text2Room fail in generating specific stylized scenes. Compared with Text2Room and LucidDreamer, our method shows less artifacts and better geometry consistency.  }
    \label{fig:compare}
\end{figure*}

\subsection{Rendering with Refined 3D Gaussian Splatting}\label{3.4}
We finally use 3D point cloud as an accurate 3D representation. The generated global-level point cloud $\Omega$ serves as the initial Structure from Motion (SfM) points, which helps accelerate the convergence of the training. First, based on the base camera poses set $P_B$ and the corresponding projected image set $I_B$, the set of moving camera poses $P_M$ and the corresponding projected image set $I_M$, we conduct the initial densification process of the 3D Gaussian Splatting and the initial 3D Gaussians $G_0$. Although the initial densification process is able to fill in the missing hole of the 3D point cloud, there is misalignment in these regions. Therefore, we then use the supplementary camera set $P_S$, the supplementary image set $I_S$ and the supplementary mask set $M_S$ to refine this process. The supplementary images and poses provide additional supervision in the second stage of training until we get the final rectified 3D Gaussians $G_1$ with global-level consistent geometric consistency with high-quality details.

\setlength{\tabcolsep}{3pt} 
\begin{table*}[t!]\footnotesize
\centering
\caption{Comparison with other scene generation methods}
\label{tab:compare}
\begin{tabular}{@{}lccccc@{}}
\toprule
             & CLIP-T score ↑  & Q-Align-Quality ↑ & Q-Align-Aesthetic ↑ & NIQE ↓ & BRISQUE↓  \\ \midrule
Text2Room~\cite{hoellein2023text2room}    & 0.291    & 3.078   & 3.089       &  5.872    & 40.85 \\
LucidDreamer~\cite{chung2023luciddreamer} &  0.296     & 3.051   & 3.035    &  6.132  & 45.78   \\
Ours         &  \textbf{0.311} &  \textbf{3.112}  &  \textbf{3.129}  & \textbf{5.025}     &  \textbf{37.97}          \\ \bottomrule
\end{tabular}
\end{table*}

\setlength{\tabcolsep}{3pt} 
\begin{table*}[ht!]\footnotesize
\centering
\caption{Ablation Study on rendered image quality.}
\label{tab:example_table}
\begin{tabular}{@{}lccccc@{}}
\toprule
             & CLIP-T score ↑ & Q-Align-Quality ↑& Q-Align-Aesthetic ↑& NIQE ↓& BRISQUE↓ \\ \midrule
\textit{w/o }supplementary cameras    &  0.295  & 3.011  & 3.002   &  5.568   &  45.68    \\
\textit{w/o} depth alignment &   0.302   &  3.035   &  3.028    &  5.156  &  40.39      \\
Ours         &  \textbf{0.311 }    & \textbf{3.112}  &  \textbf{3.129}   &   \textbf{5.025} & \textbf{37.97}         \\ \bottomrule
\end{tabular}
\label{fig:ablation}
\end{table*}

\begin{figure*}[t!]
    \centering
    \includegraphics[width=0.95\linewidth]{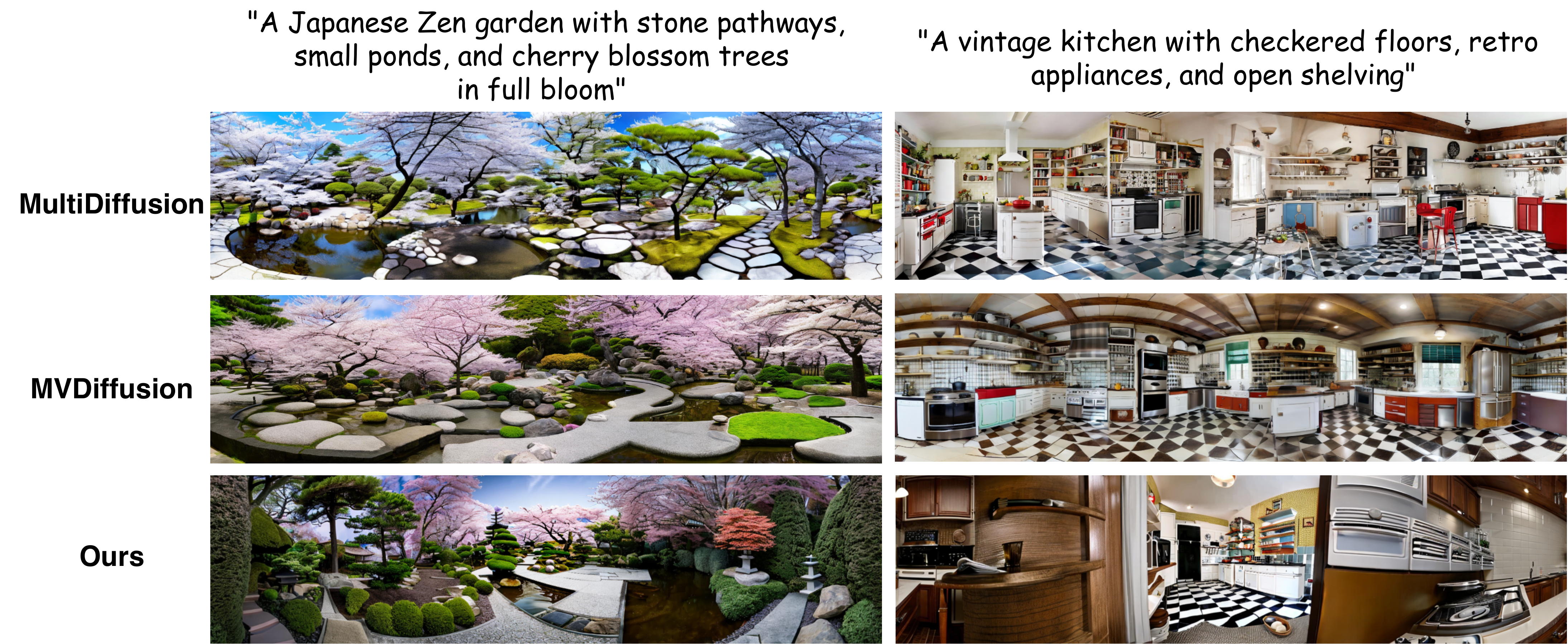}
    \caption{Comparison of Text-to-Panorama Generation. As panoramas generated by MultiDiffusion~\citep{bar2023multidiffusion} and MVDiffusion~\citep{Tang2023mvdiffusion} have both limited vertical FoV, for comparison, we only show our panorama before outpainting. Compared with previous methods, our method shows less duplicated objects and better generation quality.}
    \label{fig:pano}
\end{figure*}

\begin{figure*}[t!]
    \centering
    \includegraphics[width=0.95\linewidth]{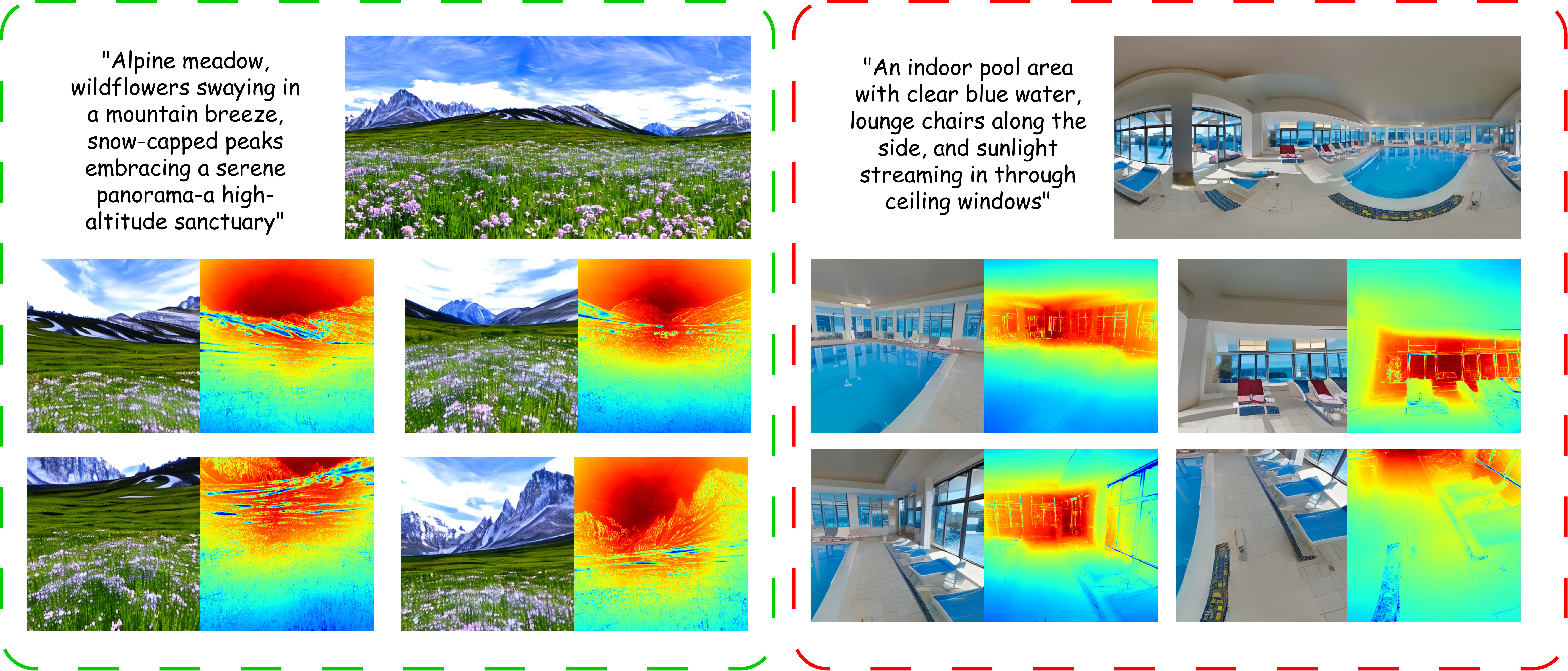}
    \caption{Our rendered rendered images with corresponding rendered depth.}
    \label{fig:depth}
\end{figure*}

\section{Experiments}
In this section, we conduct experiments to evaluate our model's performance on  both text-to-panorama generation and panoramic scene generation.

\subsection{Experimental Setup}
\paragraph{\textbf{Evaluation Metrics}}
For quantitative evaluation, to evaluate the non-reference quality of the rendered images in the scene, we render images using supplementary cameras, and employ traditional no-reference image quality assessment metrics: Natural Image Quality Evaluator (NIQE)~\citep{6353522} and Blind/Referenceless Image Spatial Quality Evaluator (BRISQUE)~\citep{6272356}. We adopt QAlign~\citep{wu2023qalign}, the state-of-the-art method in quality assessment benchmarks to evaluate the perceptual quality of image contents, which is divided into quality and aesthetic to evaluate the image quality and aesthetic quality respectively. We also use CLIP-T~\citep{radford2021learning} score to evaluate the alignment between the text input and the generated perspective-of-view images.

\paragraph{\textbf{Data}}
For the text prompts used for text-to-3D reconstruction, we use GPT4 to generate random scene descriptions. During evaluation, we use GPT4 to generate 40 prompts, including 20 indoor scenes and 20 outdoor scenes. The quantitative results are calculated through the average score of the scenes.

\subsection{Main Results}

For evaluating the quality of 3D scene generation, we compare our approach with state-of-the-art 3D scene generation methods: Text2Room~\citep{hoellein2023text2room}, which employs an iterative mesh generation approach to represent the scene based on inpainting and monocular depth estimation, and LucidDreamer~\citep{chung2023luciddreamer}, which utilizes a warp-and-refine strategy to iteratively generate point clouds for novel views and subsequently employs 3D Gaussian Splatting (3D-GS) to obtain the Gaussians of the scene. Since LucidDreamer cannot directly generate 3D-GS from text prompts, we use Stable Diffusion v2.1~\citep{rombach2022high} to generate the initial conditioning image, ensuring consistency with our method. The comparison results are shown in Fig.~\ref{fig:compare} and Table~\ref{tab:compare}.

The results indicate that Text2Room struggles to generate coherent scenes when style descriptions are included. Due to its render-refine-repeat scheme, Text2Room encounters alignment issues when there is significant variation between the generated images, which prevents the model from effectively distinguishing overlapping regions. This issue is particularly pronounced when the prompt contains numerous object descriptions. LucidDreamer, on the other hand, can only generate coherent scenes with limited camera movement. Due to the accumulation of geometry errors inherent to its warp-and-inpaint generation scheme, both Text2Room and LucidDreamer fail to maintain consistency between views, particularly during large camera movements. Consequently, these methods exhibit blurry boundaries and artifacts at the intersections between adjacent objects.
Our method produces high-quality results with smooth boundaries and fewer artifacts in both indoor and outdoor scenes. Furthermore, our model achieves robust geometric consistency even under large camera movements, setting it apart from the compared approaches.

We present results for text-to-panorama generation compared with prior methods~\citep{Tang2023mvdiffusion,bar2023multidiffusion} in Fig.~\ref{fig:pano}. MultiDiffusion\citep{bar2023multidiffusion} directly generates panoramas using rectified diffusion, whereas MVDiffusion~\citep{Tang2023mvdiffusion} first generates perspective-view images using diffusion models and then composes them into a panorama. The results demonstrate that with LLM guidance, our model effectively avoids generating duplicate objects and significantly enhances content diversity and generation quality.

We also visualize qualitative results in Fig.~\ref{fig:depth}. The results show that the rendered images show accurate depth maps, which validates the accurate geometry of our rendered results.

\subsection{Ablation Study}

We conduct ablation studies to verify the supplementary camera set and the depth scale component. As shown in Table.~\ref{fig:ablation}, incorporating supplementary cameras and semantic-preserved generative warping enhances the refinement stage of 3D Gaussian Splatting, while also reducing artifacts in the rendered results. Removing the depth alignment module results in blending issues between scenes, causing pixel misalignment and increasing geometric deviations during 3D Gaussian generation. Since 3D-GS relies heavily on accurate point cloud initialization, incorporating depth alignment reduces the misalignment between the panorama scene and the moving scenes, ultimately improving the quality of the rendered images.

We also compare the render quality of with other methods in Table~\ref{tab:example_table}. The results demonstrate that the exclusion of either the supplementary camera refinement or depth alignment leads to a significant degradation in rendering quality. These findings underscore the importance of both components in achieving high-quality scene reconstruction.

\section{Conclusion}

We proposed PanoDreamer, a text to 360-degree scene generation framework. The core insight of our method is to decompose scene generation into two phases: single-viewpoint scene generation and scene extension via moving camera simulation. The first phase uses an LLM to guide the synthesis of perspective images, which are then fused to form a panorama. During the second phase, the model is extended and improved using two different generation strategies. Our approach results in high-quality, geometry-consistent scenes, represented in the 3D-GS framework, and enables users to navigate freely along customized trajectories outside the initial, significantly broadening the range of potential applications. Our approach consistently outperforms strong baselines across a broad set of metrics.
A key challenge that we plan to explore in future work is the accumulation of error as the scene scale becomes larger.

{
    \small
    \bibliographystyle{ieeenat_fullname}
    \bibliography{main}
}



\begin{center}

\end{center}
\clearpage

\newcommand{\appendixhead}%
{\centering\textbf{\huge Appendix}
\vspace{0.5in}}
\twocolumn[\appendixhead]

\appendix

\section{Experiment Details}

During rendering, as we use pinehole cameras, reducing the camera Field-of-View(FoV) will reduce the distortions. During the projection, we set the camera FOV of the base camera and the supplementary cameras to be $60^\circ$. We set the number of base camera to be 80, with each base camera corresponding to 4 supplementary cameras. The resolution of the projected images from base cameras and the supplementary cameras are set to be $512 \times 512$.

During the two-stage 3D gaussian splatting process, for the first 5000 iterations, we use the base set to initialize the 3D Gaussians, and after 5000 iterations, we add the refined supplementary set for the second-stage refinement of the 3D Gaussians. We report the performance of the rendered results on training 10,000 iterations.

\section{Further Qualitative Results of Scene Generation}
\begin{figure}[t!]
    \centering
    \includegraphics[width=\linewidth]{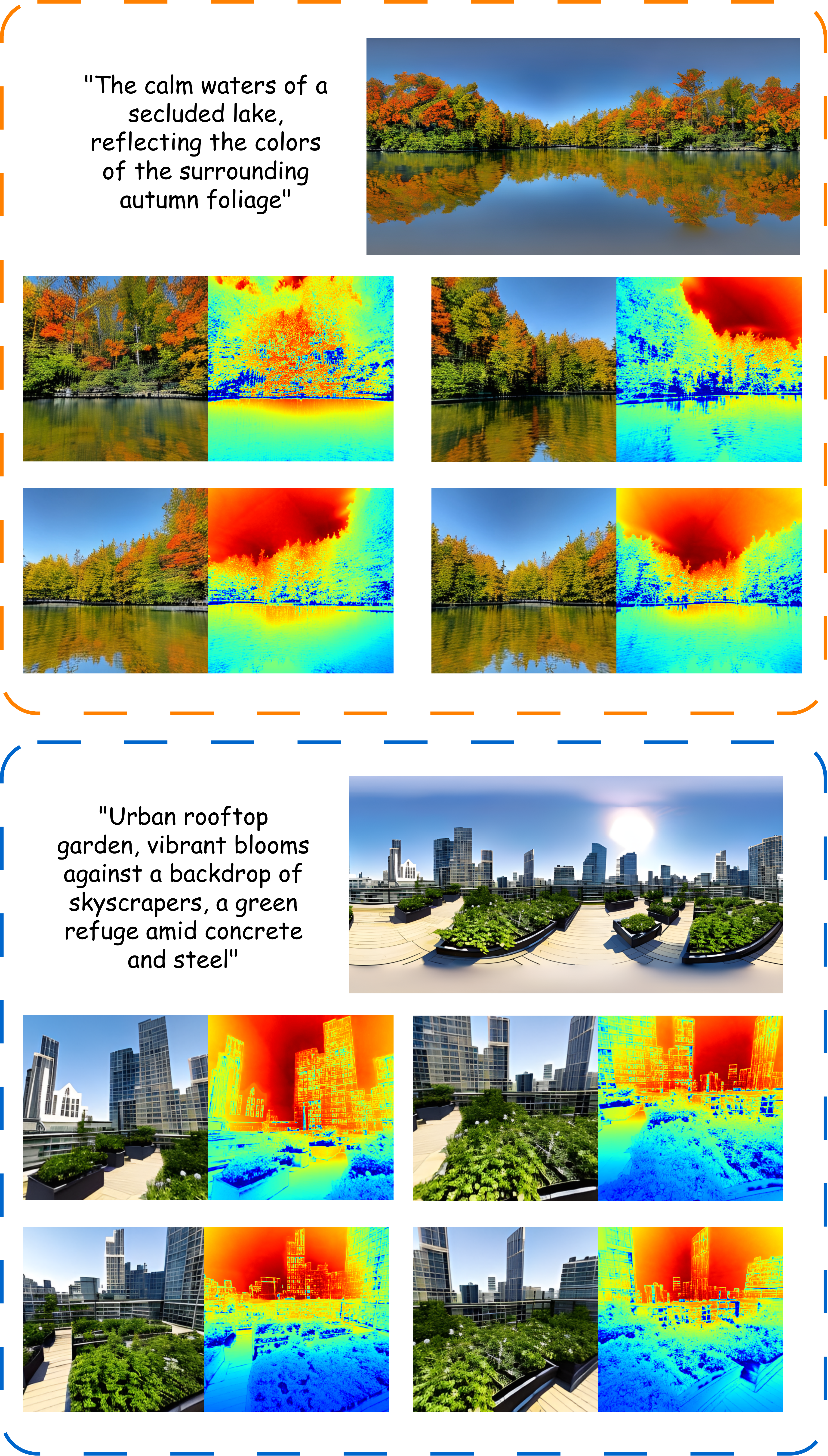}
    \caption{Additional results about scene generation. We show both the rendered images and rendered depth.}
    \label{fig:scene_supp}
\end{figure}

We show more qualitative results of our method on some scenes, shown in Fig.~\ref{fig:scene_supp}. Results show that our method not only generate high-quality rendered images, but also maintains accurate geometry and scene consistency.

\section{Further Qualitative Results on Text-to-Panorama Generation}

\begin{figure*}[t!]
    \centering
    \includegraphics[width=0.9\linewidth]{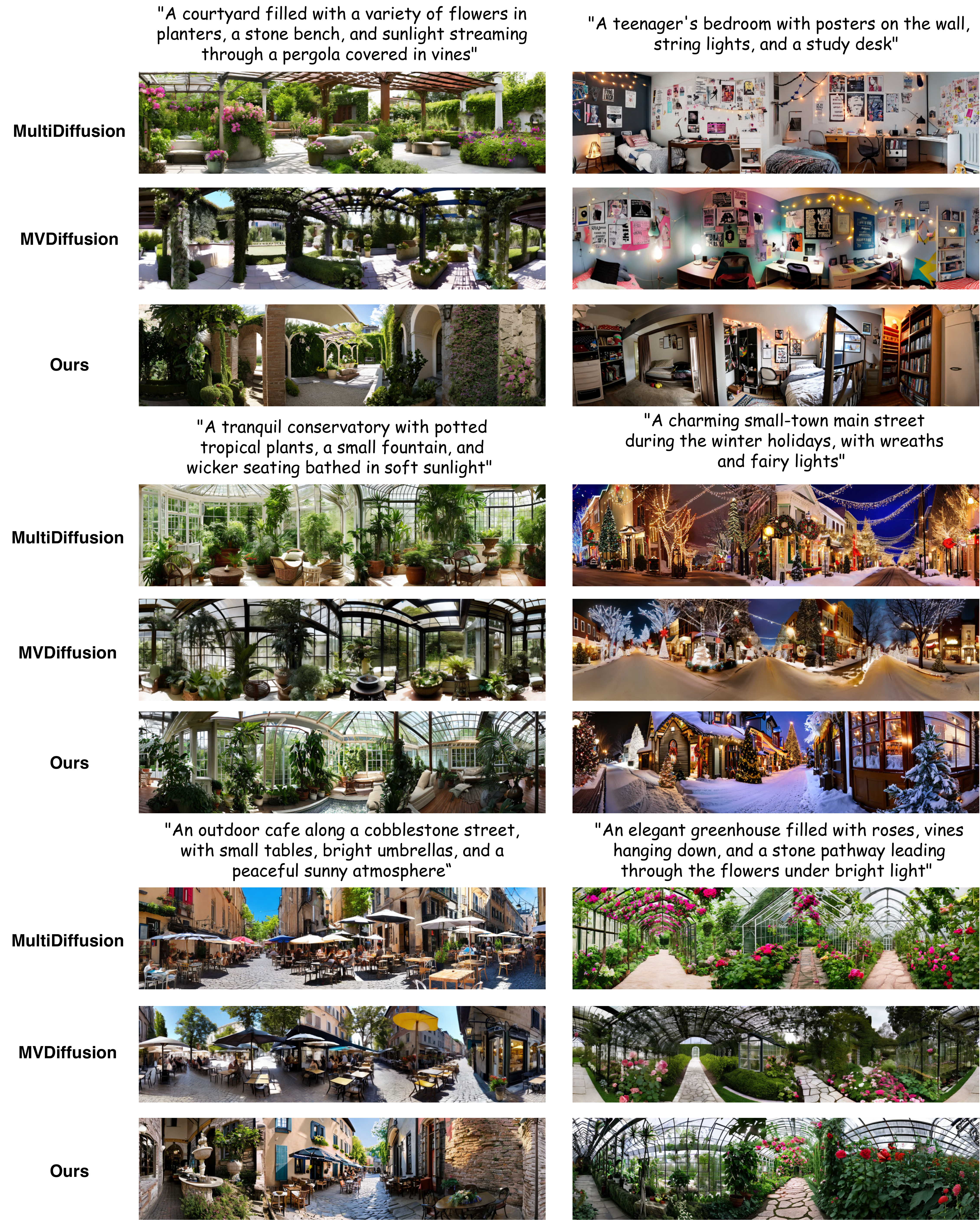}
    \caption{Additional comparison of text-to-panorama generation. As panoramas generated by MultiDiffusion~\citep{bar2023multidiffusion} and MVDiffusion~\citep{Tang2023mvdiffusion} have both limited vertical FoV, for a fair comparison, we only show our panorama before outpainting.}
    \label{fig:pano_supp}
\end{figure*}

We present more results for text-to-panorama generation in comparison with prior methods~\citep{Tang2023mvdiffusion,bar2023multidiffusion} in Fig.~\ref{fig:pano_supp}. Compared with previous methods, with LLM guidance, our method shows less duplicated objects and better generation quality.

\end{document}